\documentclass{article}

\usepackage{arxiv}

\usepackage[utf8]{inputenc} 
\usepackage[T1]{fontenc}    
\usepackage{hyperref}       
\usepackage{url}            
\usepackage{booktabs}       
\usepackage{amsfonts}       
\usepackage{nicefrac}       
\usepackage{microtype}      
\usepackage{lipsum}
\usepackage{graphicx}
\usepackage{authblk}
\usepackage{float}
\usepackage[compact]{titlesec}
\title{Isthmus: Secure, Scalable, Real-time and Robust Machine Learning Platform for Healthcare}

\author[1]{Akshay Arora}
\author[1]{Arun Nethi}
\author[1]{Priyanka Kharat}
\author[1]{Vency Verghese}
\author[2]{Grant Jenkins}
\author[2]{Steve Miff}
\author[2]{Vikas Chowdhry}
\author[2]{Xiao Wang}
\affil[] {Parkland Center for Clinical Innovation}
\affil [] { Dallas, TX 75247, USA}

\begin{document}
\maketitle

\begin{abstract}
In recent times, machine learning (ML) and artificial intelligence (AI) based systems have evolved and scaled across different industries such as finance, retail, insurance, energy utilities, etc. Among other things, they have been used to predict patterns of customer behavior, to generate pricing models, and to predict the return on investments. But the successes in deploying machine learning models at scale in those industries have not translated into the healthcare setting. There are multiple reasons why integrating ML models into healthcare has not been widely successful, but from a technical perspective, general-purpose commercial machine learning platforms are not a good fit for healthcare due to complexities in handling data quality issues, mandates to demonstrate clinical relevance, and a lack of ability to monitor performance in a highly regulated environment with stringent security and privacy needs. In this paper, we describe \textit{Isthmus}, a turnkey, cloud-based platform which addresses the challenges above and reduces time to market for operationalizing ML/AI in healthcare. Towards the end, we describe three case studies which shed light on \textit{Isthmus}’ capabilities. These include (1) supporting an end-to-end lifecycle of a model which predicts trauma survivability at hospital trauma centers, (2) bringing in and harmonizing data from disparate sources to create a community data platform for inferring population as well as patient level insights for Social Determinants of Health (SDoH), and (3) ingesting live-streaming data from various IoT sensors to build models, which can leverage real-time and longitudinal information to make advanced time-sensitive predictions. 
\end{abstract}


\section{Introduction}
Scalable ML/AI driven solutions that are tightly integrated in the clinical decision support system  \cite{jameson2015personalmeds} have the potential to improve outcomes, reduce cost, and increase efficiency in healthcare. While healthcare has made steady progress through the years in using data to make evidence-based clinical decisions through randomized clinical trials  \cite{Sterne2009imputation}, ], it has lagged to adopt ML/AI driven solutions at scale. Developments over the last decade have made use of Electronic Health Records (EHRs) widespread in healthcare organizations, which has resulted in an abundance of structured and unstructured clinical and non-clinical data  \cite{Bates2014highcost},but the industry has struggled to generate value from this data. The technical reasons for this slow uptake include uneven data quality, privacy, and confidentiality issues in building workflows, and a lack of interoperability between different systems \cite{Lucas2019Association}. To counter the concerns above and deploy ML predictive models in the clinical workflows, we developed a turnkey ML platform called \textit{Isthmus} to seamlessly develop, test, deploy, evaluate, and retrain predictive models.  

Developed under clinicians’ guidance and oversight from Parkland Health \& Hospital System’s stakeholders, \textit{Isthmus} balances the domain-rich customization needed for building clinical, predictive models in healthcare with the need of having a turnkey, secure, HIPAA compliant solution. Consequently, the platform reduces time to market to integrate and update AI/ML models to improve the quality and safety of a patient’s hospital experience and drive costs down by impacting both clinical and operational outcomes.

\section{Methodologies}

\subsection{\textit{Isthmus} Framework}

Our vision was to build an end-to-end machine learning framework which would make predictive model development, deployment, evaluation, and retraining seamless by reducing the time to market for integrating clinical predictive insights in clinical workflows to make them actionable. To achieve the goals above we built a cloud hosted, machine learning platform called “\textit{Isthmus}.”  

\textit{Isthmus}’ end-to-end workflow will reduce implementation time when there are multiple models  developed and deployed on it,by reusing the environment toolsets, technologies, and data/feature engineering pipelines to either train or score using an already deployed pre-trained model.   

\begin{figure}[H]
\centering
\includegraphics[width=\linewidth]{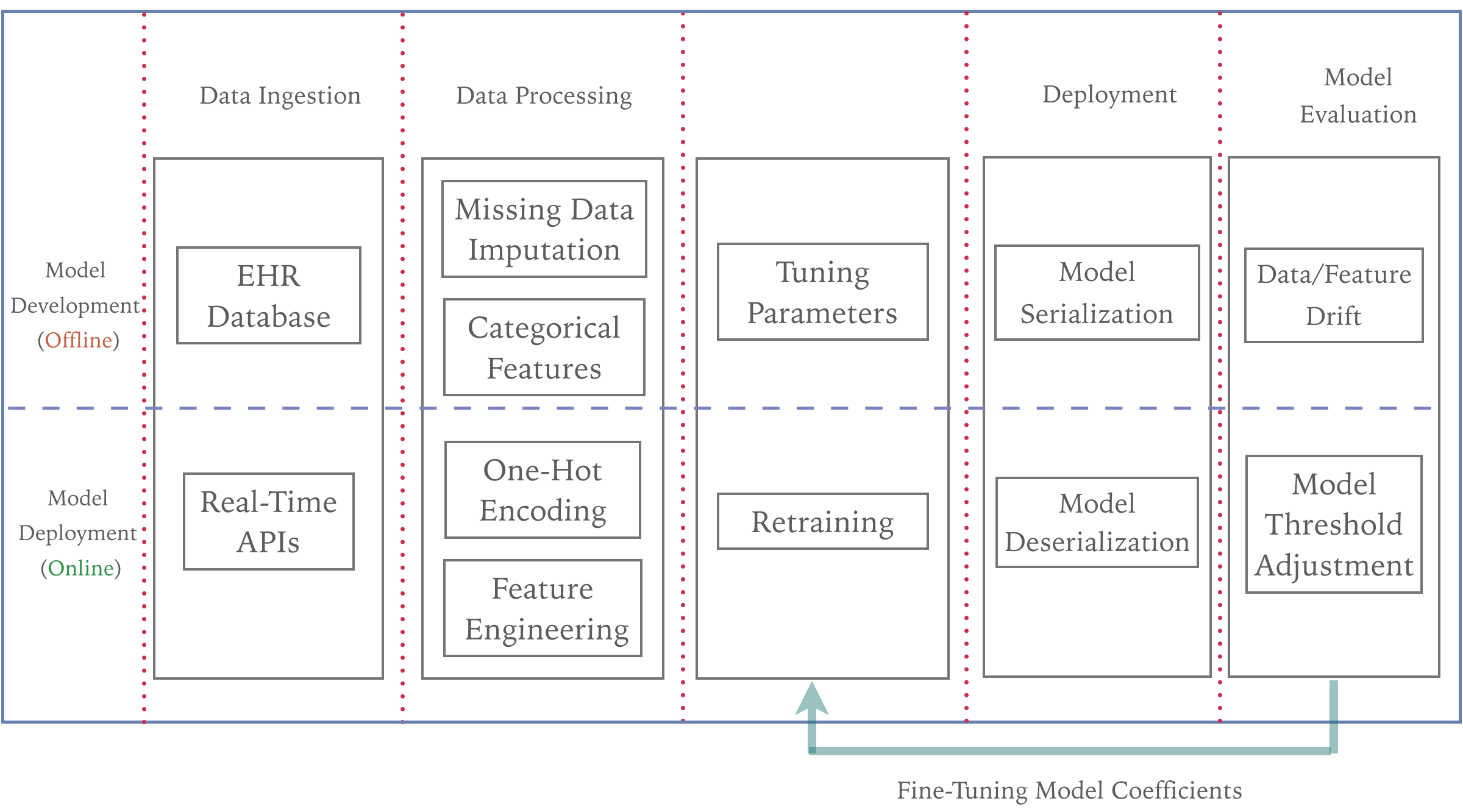}
  \caption{\textit{Isthmus} Framework: End-to-end cloud hosted machine learning platform for healthcare}\
\end{figure}

 Part of the motivation to build such a flexible but scalable and configurable framework was due to the curated set of data transformation techniques that data scientists must perform in terms of imputation, categorical encoding of continuous variables or aggregation of healthcare datasets before building them into features to train a predictive model \cite{Alvin2018Scalable}. Our framework will reuse these techniques as packages, which can be invoked in the deployment flow so that there is consistency in the way features are created for model training and model scoring. Thus, there is standardization of the training and deployment/scoring workflow, which helps in quickly learning through prospective testing the key components which can trigger data or feature drifts as the model runs in a real environment.  

 Doing this in the same controlled environment, which can ingest either historical or real-time data through the same APIs or secure connection, is key to productionize this within healthcare providers’ and payers’ systems to drive value. To achieve this, we hosted the entire framework in a secure HIPAA compliant cloud infrastructure to deploy as a turn-key solution as outlined below.

\subsection{Cloud Hosted Secure Healthcare Specific Turnkey Platform}

\textit{Isthmus} is hosted on a cloud-based infrastructure (Microsoft Azure Cloud Platform \textsuperscript{TM})\cite{microsoft}. This infrastructure is set up with all the state-of-the-art functionalities, including network security, data replication, disaster recovery, and fault tolerance necessary for a robust and enterprise-grade software-as-a-service (SaaS). Cloud resources – compute and storage, both leverage economies of scale to keep the technology costs at a realistic level without having a need to maintain a large number of on-prem resources. Thus, being cost-effective as well as scalable and configurable, \textit{Isthmus} can be adopted by healthcare organizations and systems of all sizes. 

\subsubsection {Data Engineering Pipeline}

Healthcare data by its very nature is highly complex, high dimensional. and of inconsistent quality. For this data to be useful, it needs a systematic data ingestion approach to collect, store and integrate data-driven insights into clinical and operational processes. To quickly ingest this multi-dimensional data and scale, we developed a configurable and flexible data ingestion pipeline solution, which enables the ingestion of all the relevant health data, including clinical data, claims data, Social Determinants of Health, and streaming IoT data. The data ingestion pipeline is also ready to allow the ingestion of genomics data and high-quality diagnostic imaging data as it becomes available.   

The data ingestion pipeline is based on a simplified architecture (Figure 2), which enables user defined transformations for real-time scoring, cleaning, and de-duplication without requiring additional middleware. The raw data is pulled via RESTful API calls to the EHR’s API servers or through regular intervals of data fetch using secure file transfer process. Generally, these API servers are the hub for all the API requests, which facilitates the connection between the EHR organizational users and the operational database management system to stream near real-time data seamlessly as a JSON response through the web service APIs upon service request.

\begin{figure}[H]
\centering
\includegraphics[width=\linewidth]{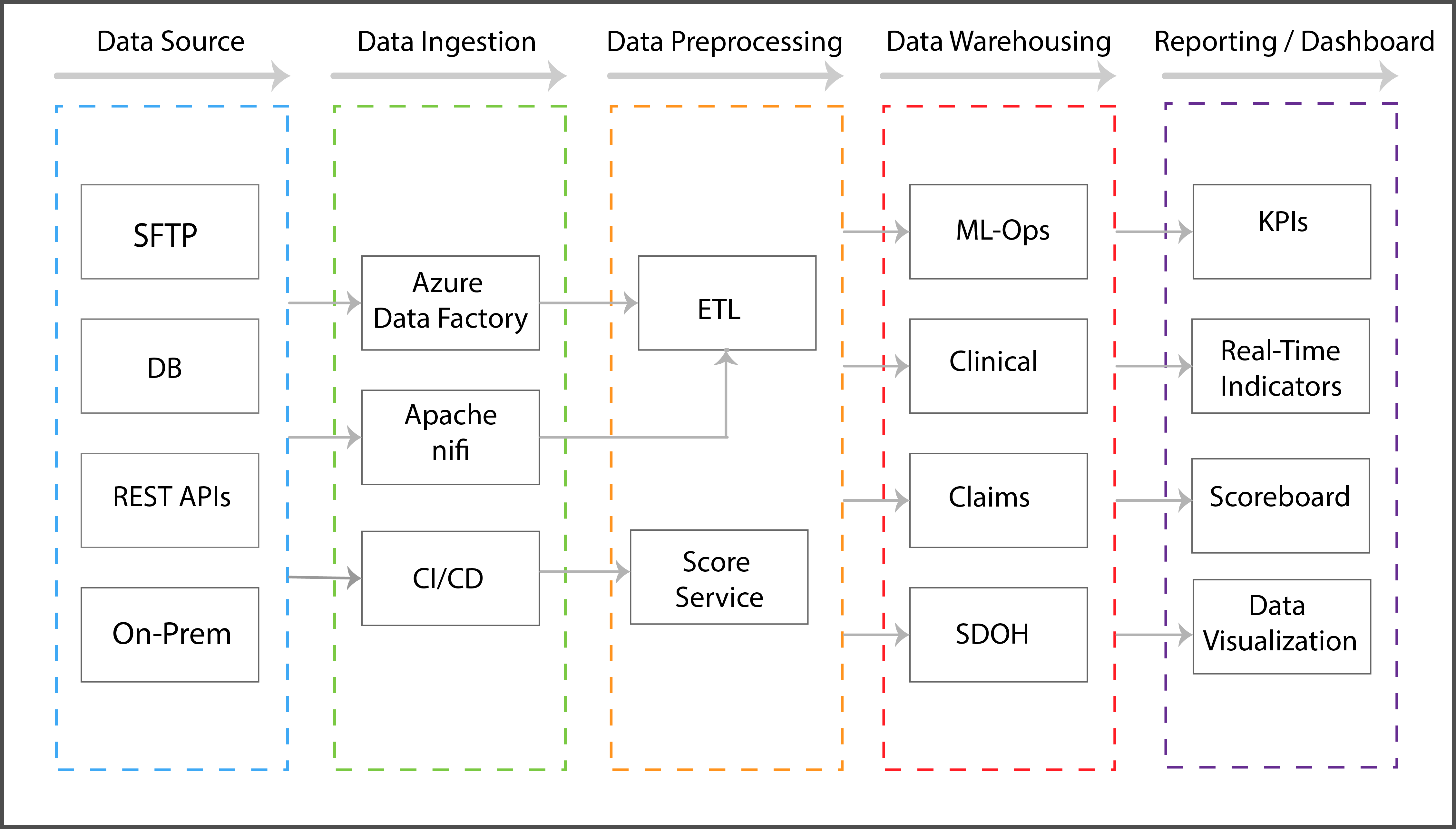}
  \caption{Data Orchestration Engine}\
\end{figure}

The clinical data is mainly comprised of patient demographics, encounter related information, laboratory test result components, medication orders and laboratory vitals. These are pulled through APIs and are parsed through a configurable data munging script where the JSON objects are transformed using user defined templates that are specifically designed for each API response and are aggregated into patient level JSON responses. The pipeline is fully automated, and it ingests the patient level data in batch mode, where the batch size is based on SLA requirements. Thus, the pipeline is scheduled to trigger based on SLA requirements and it continuously pulls the data from the APIs and performs the desired transformation and filtering operations. 

This patient level raw JSON data is then preprocessed using an imputation and filter logic, which transforms this data into clinically relevant features that are fed to the machine learning models using a scoring logic script to predict the risk of the acute care condition based on the pre-trained model.The scoring script generates the score response which encompasses the transformed features and the identified risk levels associated with the patient. These responses are aggregated in batch mode, and after cleaning, they are converted into SQL tables using a database operation script and ingressed into the PostgreSQL database. The data is stored in a very secure and reliable manner within the PostgreSQL database. The raw JSON responses are pushed to the Azure Data Lake\textsuperscript{TM} to preserve the raw patient level information for audit purposes. 

\subsubsection{ Configurability, Extensibility and Experimentation}

 Predictive models in the healthcare industry are discrete in nature, especially for acute care systems \cite{Bret2009Rethinking}, where we observe a high complexity of imputation logic and disparate features. The rest of the workflow has unified commonalities for all the models with identical evaluation patterns. If the configuration of these models is distributed and model-specific, it will add an overhead of maintaining and versioning multiple configurations for a given model as well as different environments. This will also increase the cost of additional infrastructure to manage different models. 
 
The \textit{Isthmus} platform provides a unique way of deploying and executing a model workflow for scoring using a single code base, which can support multiple models and versions using a single configuration file as shown in figure 3. It is also designed to use a single infrastructure cluster to execute any number of scoring workflow pipelines in parallel and automate the scoring process using Continuous Integration and Continuous Delivery processes (CI/CD). The use of the configuration methodology facilitates easy upgrading to an existing model or serving a new model in the pipeline workflow as it has a very short delivery cycle. 
\begin{figure} [H]
\centering
\includegraphics[width=\linewidth]{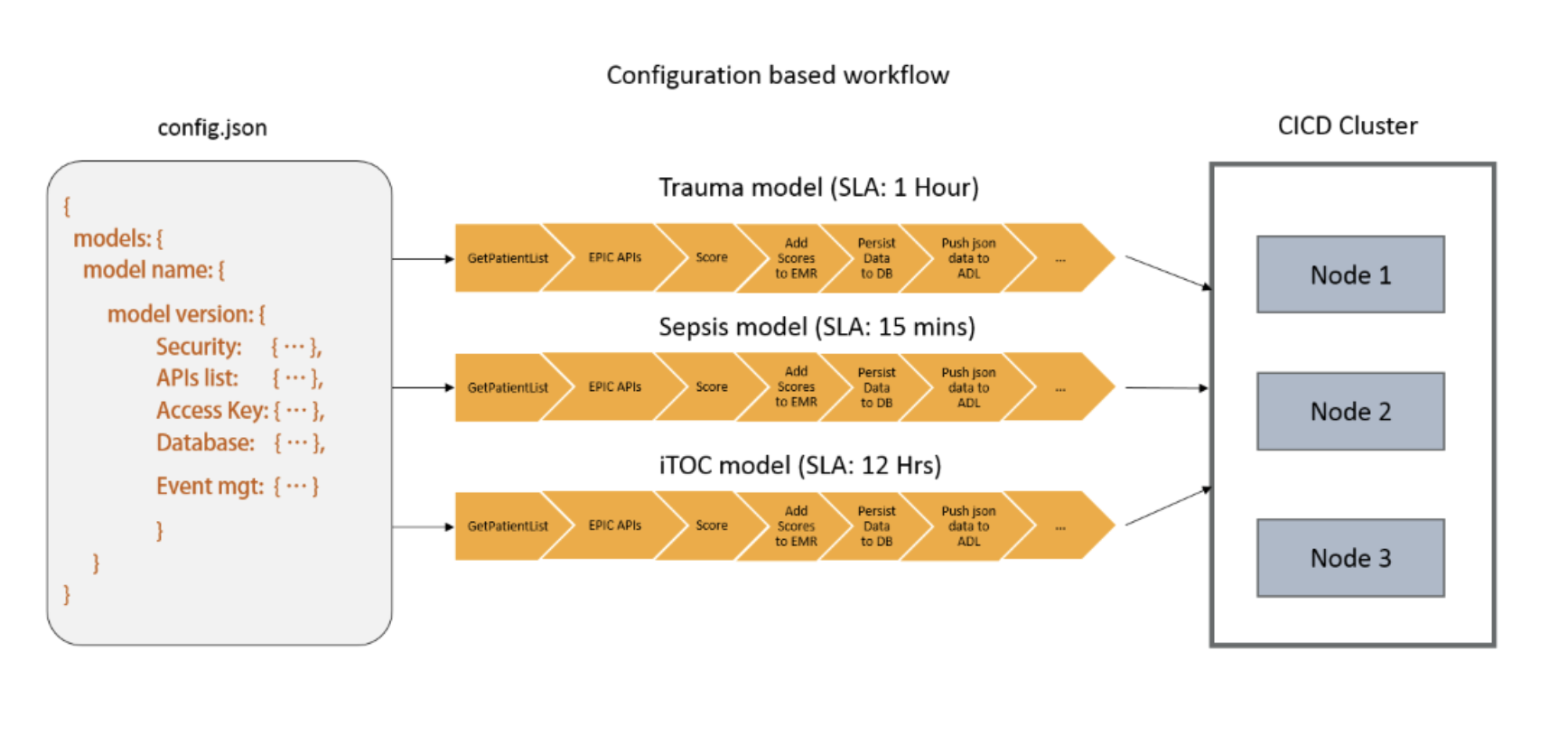}
  \caption{Configuration based workflow}
 
\end{figure}
\subsubsection{Security and governance}

All data collected in the context of healthcare is subject to specific privacy and regulatory requirements. Depending upon the type of data, and the context of the data collection and use, it can be Personally Identifiable Information (PII) or Protected Health Information (PHI).  This information is often the target of security breaches. This is another reason why every access, analysis, and appending (if any) of this data needs to be audited, monitored, and documented. In the US, compliance with the HIPAA privacy law was mandated among all entities which either create or access PHI. This, in addition to the mandatory business continuity safeguards to avert a possibility of the health system facing outages of care services, puts additional constraint on the traditional feature and data engineering pipelines in a machine learning system. Challenges with data security through traditional on-prem systems and support of life-critical applications are solved by the \textit{Isthmus} platform by adopting a cloud-first strategy. Additionally, the following Azure\textsuperscript{TM} enabled features utilized on \textit{Isthmus} address the security concerns above. 

\textbf{Access Control}
\newline
The \textit{Isthmus} platform leverages unique cloud-based security policies, such as the Azure active directory-based service principal for access control as an identity to manage applications and hosted services on the cloud and handle sensitive information (PHI). This eliminates the need for user level login to the cloud applications. Role Based Access Control (RBAC) uses Active Directory policies for managing the authentication. \textit{Isthmus} provides a single role-based access to multi-institutional EHR data. 
 
Additionally, the platform also provides a comprehensive, immutable log management service with easy access across deployed applications using elastic search and Kibana\textsuperscript{TM} dashboard. This ensures a single point of reference to test for any application or system level logs in a responsible manner. With the help of app-insight notifications, the \textit{Isthmus} platform has provisioned real-time alerts for any configured event, including an exception in application or missing data from the source API to provide real-time alerts.  
 
\textbf{Server maintenance} 
\newline
In a traditional healthcare setting, on-premise infrastructure solutions are constricted by cost and complexity of maintaining the hardware, firewall, software licensing, and the additional overhead of patch management software and its upgrades. This integration capability is limited to automate any patch management process and to roll out the access policy in a distributed project environment. 
 
The \textit{Isthmus} platform is engineered to easily automate with a cloud-based solution in a limited build and setup time. Moreover, there are prebuilt images developed on this platform which can be customized based on the application requirements. Thus, it makes our infrastructure scalable to meet the minimum performance requirements as well as enables the cloud applications to be easily migrated/replicated to another environment. 

\subsubsection{Data replication and fault tolerance}
 Data replication and fault tolerance with an on-premise infrastructure needs enormous upfront investment. With an increase in processing and data scaling based on utilization, it becomes challenging to manage the business continuity needs. A distributed architecture application like a prediction model has serious challenges related to a single point of failure, which makes it substantially difficult to manage the continuous workflow without interruption when one or more of its components fail. This entails the need of the fault tolerant system design to prevent disruptions due to any single point of failure. 
 \begin{figure}[H]
\centering
\includegraphics[width=\linewidth ]{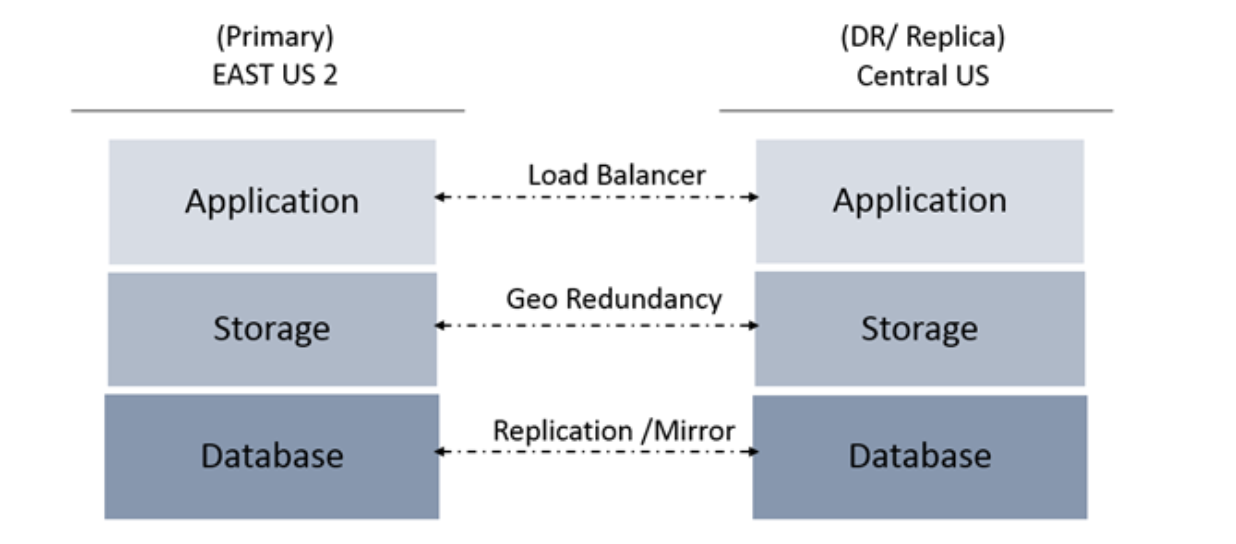}
  \caption{Disaster Recovery and Fault tolerance}
\end{figure}
The \textit{Isthmus} application is engineered to overcome these shortcomings and has the capabilities to scale up and accelerate the prediction model workloads to meet the needs of high-performance computing, low-latency, high-bandwidth network communication, and memory-intensive requirements. We adapted this cloud-based solution to resolve the lift and shift the problems like infrastructure upgrade, scalability, transfer and deployment at multiple locations using automated processes and containerization. This has considerably reduced the cost of infrastructure and engendered flexibility for migration/deployment on the cloud environments with minimal application level changes for the code, database, and data model architecture.  
 
\textit{Isthmus} applications have been developed with well-defined replication graphs and disaster recovery strategies for their database and support systems by imposing identical servers running in parallel replication with a mirrored backup of database and system level logs to ensure high levels of data availability. These applications are designed using the micro-services based architecture to reduce the redundancies from all the key components by performing similar activities in each workflow. 

\subsubsection{Log management and alerting system}
Logging is an essential component for any automated system as it serves as a flashlight on the black box module which is being deployed on a disturbed cluster. This logging information is generated in real-time on the \textit{Isthmus} platform and helps in validating the stability of the system through warning and debug logs. This log data is fed to a high scale analytical engine (elastic search), which is a full-text search engine. This data is then neatly integrated with a visualization dashboard like Kibana\textsuperscript{TM} and provides feeds to self-hosted web front applications using restful APIs. This visualization provides monitors and performance metrics based on application level logs of the automated pipeline for Predictive and analytical applications. This also ensures quality delivery of the model serving on this platform and a quick debugging capability for any production outage. 

 For any production environment which is automated, having a notification system is critical given the fact that no workflow/infrastructure is perfect. In addition to the log management system, a slack based notification service is also integrated with the \textit{Isthmus} platform to get real-time alerts about the production pipeline. This service has been deployed in real-time in production and is helping the engineering and the data science team to be fully aware of the live status of the pipeline and the patient risk scores. It is found to be useful to behave as the first line of action in the case of anomaly detection. The notification system captures both infrastructure and application failures/exceptions. Thus, this alerting system ensures immediate action and remediation in case of any failed events. 

\subsection{Model Evaluation}
The platform is designed to be a generic multipurpose data science engine. The flexible architecture of this platform has allowed the plug in and use of functional decision-making modules that can run asynchronously without disrupting the integrity of the system. The prediction service on the platform can be leveraged by the model evaluation service where real-time predictions can be interpreted by the models on the fly thereby making it extremely useful for the data scientists and clinicians (or stakeholders) to get actionable insights. 

\subsubsection{Model Interpretability}

The storage data lakes provisions access of historical data set while the model is running live in production.  This data set is pushed to a model explainer script. This script helps in interpreting the model predictions by extracting the top contributing features that helped in making the predictions. This has been proved to be extremely useful by clinicians for making real-time decisions. It has also allowed them to provide necessary feedback to improve the existing models or generate newer models. 

\subsubsection{ Silent Mode Testing }

Before the model goes live in production, the model is validated in a silent-mode testing environment where the model performance is evaluated by the team of data scientists and clinicians to ensure that the model gives meaningful results and is ready for integration into actual patient care workflows. The \textit{Isthmus} platform provisions this silent mode testing framework to follow all the best practices for the model deployment.  

\subsection{Model Retraining and Development }

In a real-time setting, it is always important that the model is highly stable and easily configurable. Thus, to address this problem, the platform allows retraining capabilities using the same data that was ingested into the model through APIs. The platform leverages this data and helps to generate multiple versions of the model which can be deployed by simply editing the model signature. The platform facilitates the data scientists to perform statistical tests to keep these models updated with the new upcoming data streams.  

These features make this platform unique as the same infrastructure with few modifications can be used for model deployment and model development. In addition to the functionalities mentioned above, there is a re-usability of the code and interoperability of the system which makes this platform highly scalable. Thus, this \textit{Isthmus} platform is an end-to-end system which encompasses all the software development processes into a robust function system.

\subsection{Multi-disciplinary team}
Healthcare is not the only highly regulated industry, but it is an industry with a  strong ethical, legal and moral underpinning to its regulations. Every clinical decision made either by a physician using either her judgement or the insights given to her by an ML system can have significant legal, financial, and/or life impacting consequences. Against this backdrop, there has been seminal work [8] conducted in demonstrating how individual-level data can be affected by historical prejudices and hence engender systematic biases in the decision of the ML system trained on it. Although there are many statistical methods to adjust for this unfairness, we believe that engaging clinical, operations and financial experts, and other stakeholders from the beginning is critical to ward off such biases and tune the algorithmic features to thresholds acceptable for the environment for which the model is being trained.
 \begin{figure}[H]
\centering
\includegraphics[width=\linewidth]{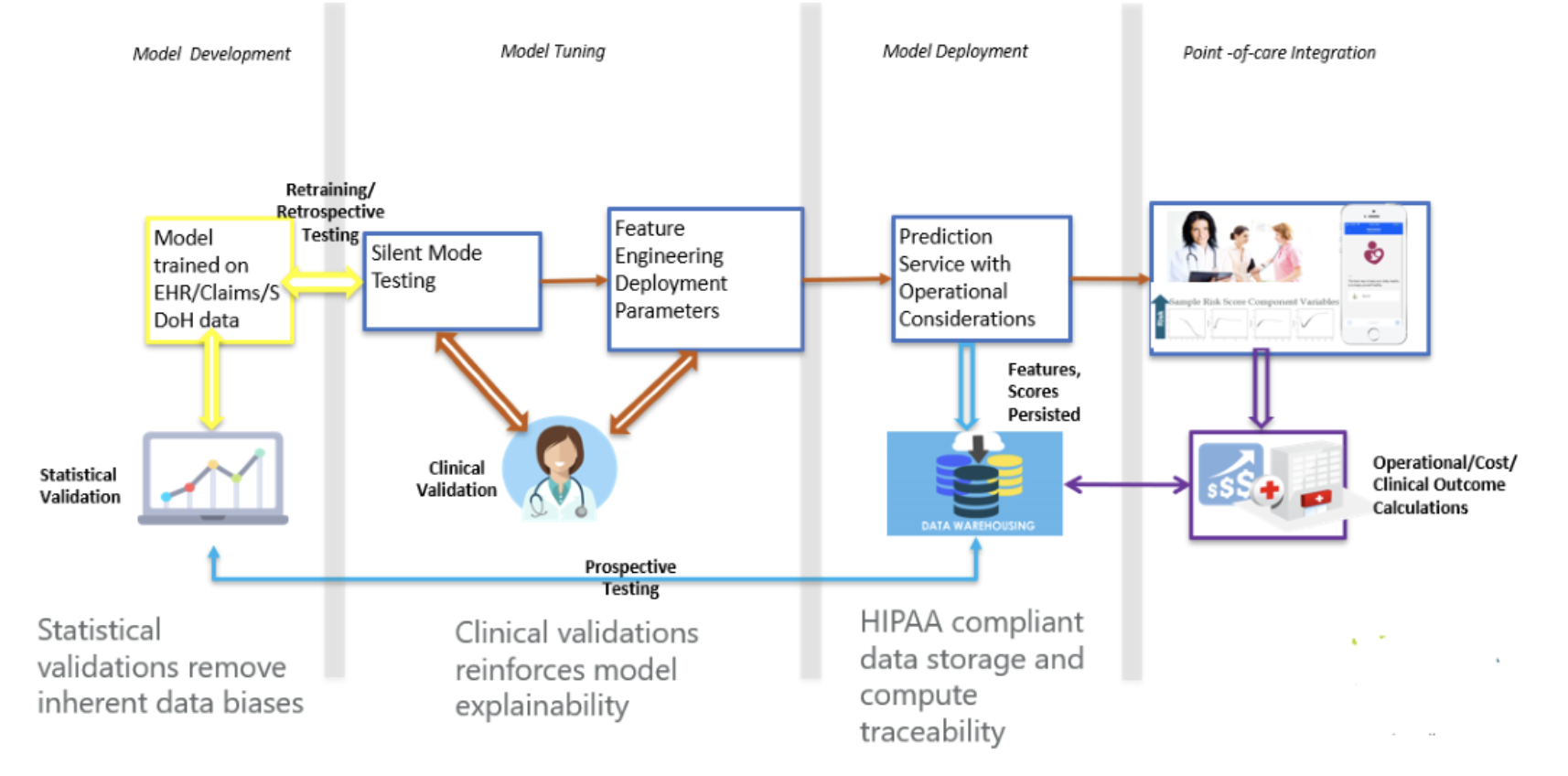}
  \caption{ML workflow with Clinical Decision Support System}
 
\end{figure}
 Involving clinicians helps in not only understanding the intimate relationship between different features, but it also gives a realistic understanding of issues related to physician cognitive overload [9] to improve user experience and help navigate actionable insights. Unlike being a fire-and-forget scenario, healthcare model development processes for clinical interventions are iterative in nature and need both retrospective as well as prospective testing [10,11]. To that effect, we adapted traditional statistical model development to a multi-faceted process, which combines pure statistical validation with clinical validation and brings in elements of clinical decision support systems. 

\section{Discussions and Results}
\subsection{Trauma case study (end-to-end framework)}
Trauma surgeons work in a high-stress world of rapid decision-making. One of the most critical decisions they must make is whether a patient is stable enough to operate on, or if the Trauma team needs to focus on providing life-saving care. Trauma surgeons must make this decision quickly and accurately, as operating could save or take the life of their patient. Traditionally, prediction of survivability in trauma patients has always been based on risk assessment algorithms like TRISS, RTS, ISS [16]. However, all these algorithms are dependent on the Abbreviated Injury Scale (AIS), an anatomical scoring system that only considers a patient’s injuries on arrival \cite{}. There is not a standard system in place to track the trauma patient’s status continuously over time, which is necessary for clinical decision-making as the patient’s condition changes while in hospital.  

In collaboration with Parkland Trauma Center Team, PCCI created a Trauma predictive model with the aim to monitor and guide clinical decision making during the first 12-72 hours of inpatient stay. This model accurately predicts mortality within the next 48 hours with 93 \% accuracy in test data. However, to impact clinical decision-making 

 PCCI’s Trauma predictive model needed a platform to bring the model to fruition with predictions made periodically every hour. Without integration into clinical workflows, PCCI’s Trauma model predictions would just add noise to provider decision-making. It was critical that the Trauma Score would be easily accessible for each patient that the trauma surgeons needed to make a timely decision on how to intervene appropriately.  

\subsubsection{\textit{Isthmus} as the Solution }

\begin{figure}
\centering
\includegraphics[width=\linewidth]{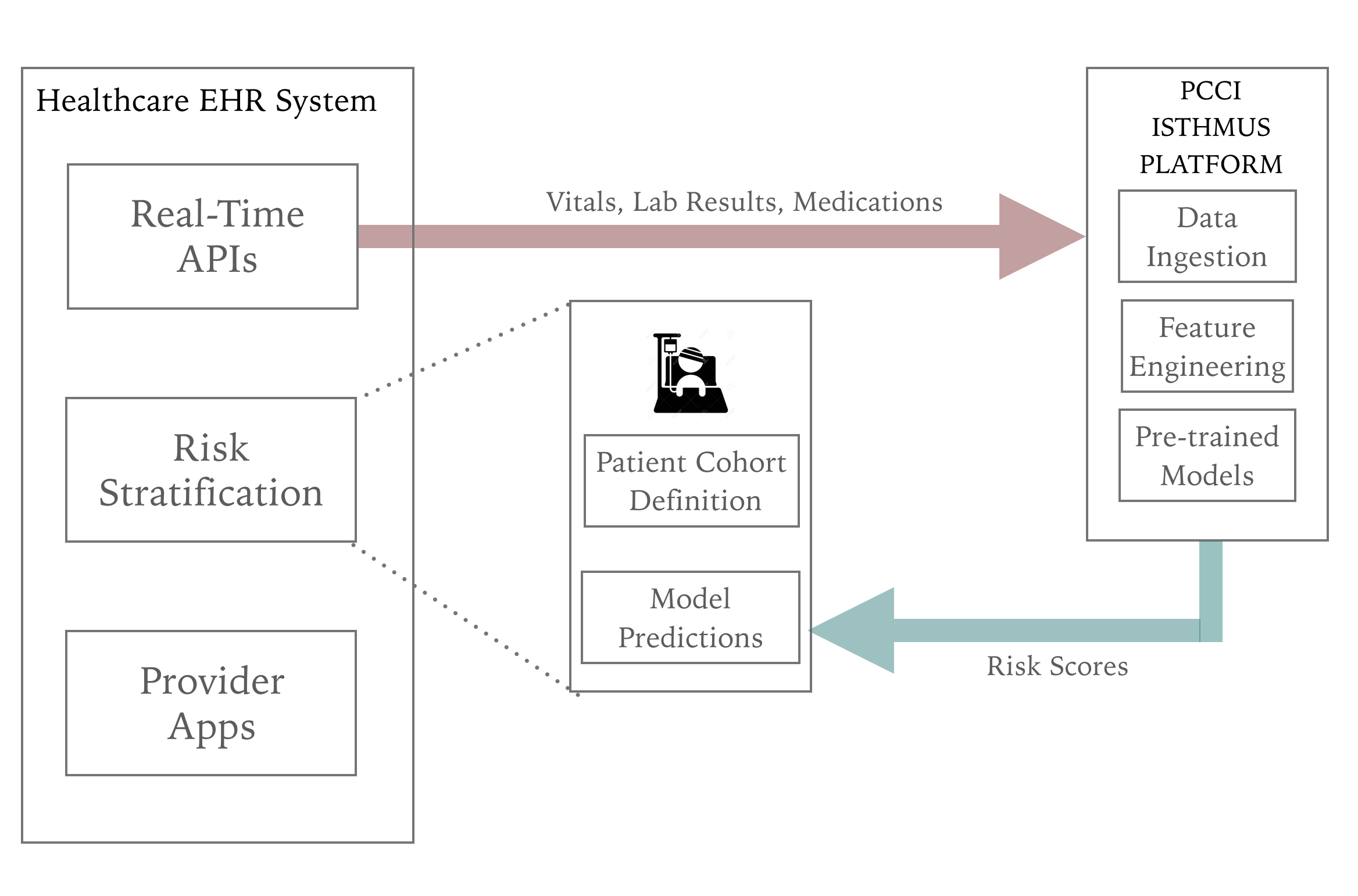}
  \caption{Trauma Model Deployment Cycle}
\end{figure}
Every hour, up-to-the-second data is pulled directly from the EMR via an API and ingested securely into the \textit{Isthmus} platform.  In less than a minute, \textit{Isthmus} cleans and transforms the API data, generates the Trauma Score, pushes  it back to the EMR, and securely persists the data and results for later analysis. Providers can then view the score directly in the patient chart. \textit{Isthmus} enables continuous, hourly predictions in the most critical window, 12-72 hours post-admission. Providers know they are always seeing a score based on up-to-date labs and vitals, so they can make decisions confidently. Using APIs and custom patient chart builds, the Trauma Score is delivered directly to the front page of the patient chart (an item called Storyboard that provides the most critical information for patient care in a column that persists as providers move through the patient chart). At a glance, providers can view and monitor the score, without a single click in the patient chart.  

 \textit{Isthmus} has several abilities that have made it a perfect launchpad for the go-live of the Trauma model. These include its (1) ability to run near real-time predictive models end-to-end via data retrieval from APIs, (2)services to perform data cleaning, data transformations, feature engineering, scoring as per the need of the model, (3) functionality to integrate with EHRs to add the scores to patient chart, (4) mechanism to securely persist the data of the scored patients for future analysis, and (4) capability to monitor/ evaluate model’s performance.

\subsubsection{Trauma Score Deployed}
The Trauma Score has been running live at Parkland Hospital for 1 month, as of the time of publication. It is currently in a clinical validation stage, with providers using the score to validate their own assumptions. So far, the Trauma Score has been generated for 134 patients, correctly predicting 4 out of 4 mortalities. Anecdotally, providers say the score matches their own predictions, formed after a thorough read of the patient chart. Providers have indicated that promising results so far mean that soon they will begin to trust the score and use it for decision-making. This will save them critical minutes, allowing them to focus on operating or life-saving care over reviewing patient charts. These critical minutes could be the difference between life and death for high-risk patients.
\footnote{CDI was supported in part through a grant from Community Council of Greater Dallas (CCGD) and was developed collaboratively with CCGD, DFW Hospital Council Foundation and University of Texas at Dallas}
\subsection{Community Data Initiative (CDI) on \textit{Isthmus} }

The healthcare industry is converging around a common realization that understanding the full context of an individual and the community where s/he lives is critical to improving outcomes and lowering costs. Factors such as transportation access, safe neighborhoods, food availability, school performance, and economic opportunity have always been part of the conversation around healthcare but now, there is an urgency to quantify their impact on the health of individuals and communities. However, data for these social determinants of health is not easily accessible. Documenting social determinants of health is typically not part of the standard documentation workflow during an individual’s clinical encounter. And zip code level indicators are not specific enough to make meaningful interventions. 

 With help from partners across the Dallas community, we utilized the \textit{Isthmus} platform to develop a block group level, longitudinal collection of about 60 key indicators that can be used to measure health, resiliency, and economic vibrancy of neighborhoods. We made these indicators available in an interactive format through a dashboard hosted on the same platform. 
 \begin{figure}[H]
\centering
\includegraphics[width=\linewidth]{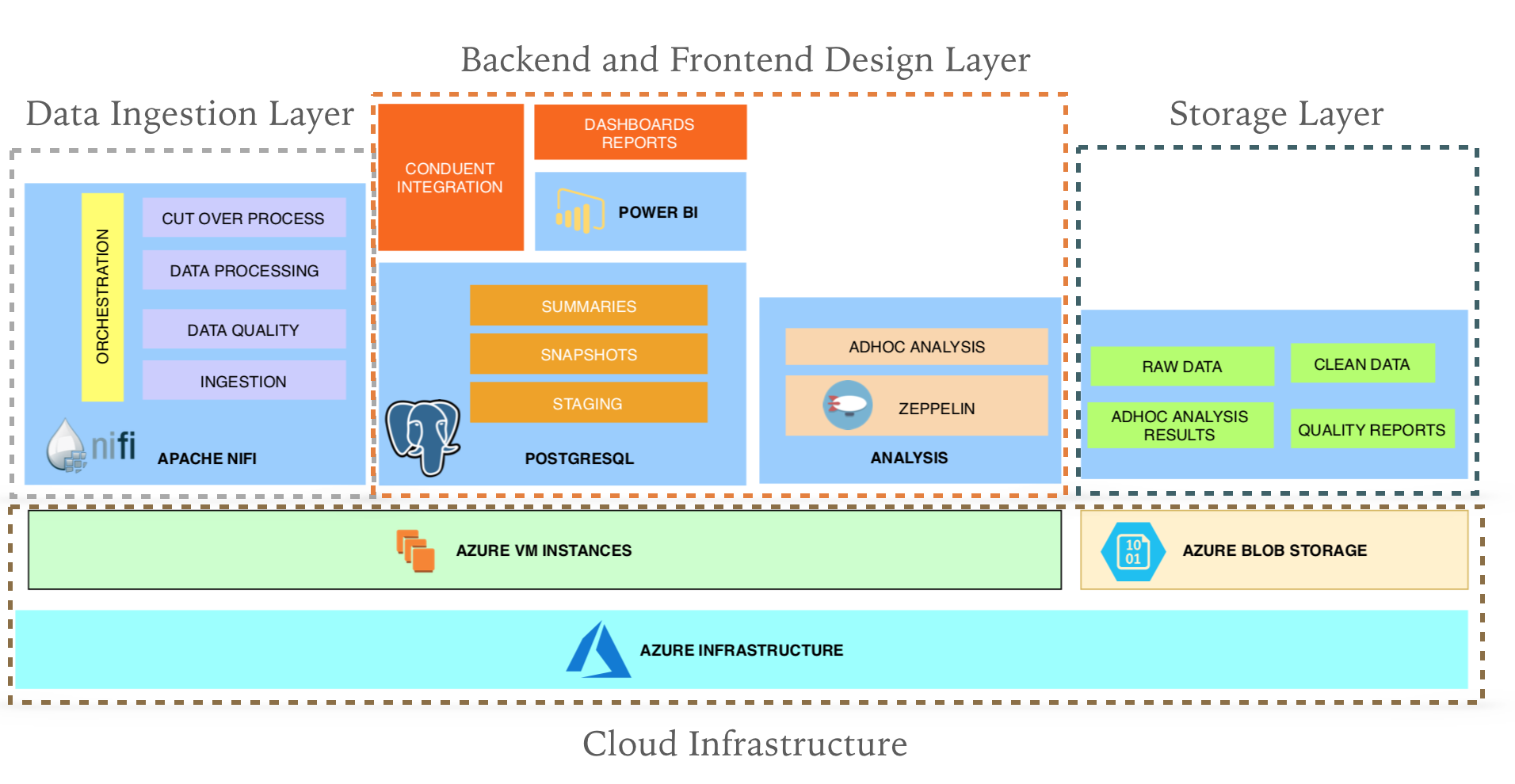}
  \caption{CWDI Infrastructure on \textit{Isthmus}}
\end{figure}

By hosting everything on a single platform, we have been able to create a streamlined process for ingesting, cleaning, analyzing, and visualizing data. Through an automated pipeline using Apache NiFi, data is ingested into the blob storage, where data is cleaned according to predefined scripts. This cleaned data is eventually ingested to the PostgreSQL database management system, which is a free and open-source database. The dashboard tool pulls data from the PostgreSQL database management system.

To quantify the impact of CDI data on specific risk conditions, we evaluated its ability to predict the risk of pre-term birth when combined with clinical and claims data. This predictive model was developed at PCCI and uses social determinants of health (SDOH) features extracted from the CDI dataset. These SDOH features proved to be significant for making predictions with this model. Thus, the data from the CDI platform is helping the data scientists and machine learning engineers demonstrate a holistic view of factors that impact preterm birth. By including the data from external sources into a model that uses clinical data, the overall model performance has improved. 

\subsection{Internet-of-Things on \textit{Isthmus}}

The use of internet of things (IoT) applications in the healthcare industry is rapidly growing. The \textit{Isthmus} platform has piloted the IoT data ingestion pipeline with indoor air quality sensors, where real-time data from these sensors’ APIs were ingested into the platform. This data was massaged and munged in batch mode using the data cleaning modules developed on the platform.

There was high reusability of the configuration which was developed for the machine learning system as the codebase was made generic to reduce redundancies across multiple projects on the platform. The IoT data is stored and maintained in a PostgreSQL database on the platform with fault tolerance and disaster recovery functionalities. The futuristic goal is to integrate this rich IoT data with the existing machine learning models to have additional features which are believed to improve the model predictions.  

\section{Conclusions}

This paper introduced the \textit{Isthmus} Platform, an end-to-end system for developing and deploying ML models. Using \textit{Isthmus}, data scientists can use their familiar ML toolkits and libraries to create models, perform statistical tests, and deploy them. 

We addressed the challenges of integrating ML models into application development and model sharing. The proposed architecture supports the sharing of pretrained models across different ML modules run-time environments. As illustrated by the case studies, \textit{Isthmus} provides project level isolation and focuses on code reusability, rather than reinventing the development pieces. The platform demonstrated versatility in terms of serving a prediction service, ingesting IoT data, and integrating an SDOH feature engine. 

 In the future, we hope to (1) integrate real-time model training with streaming data by tuning the model- coefficients on the fly and (2) perform A/B testing and multi-armed bandit testing to ensure the model stability and reliability over time.  

\section{Acknowledgments}
We are extremely grateful to Lyda Hill and the Parkland Foundation for providing financial support to design and build this platform to serve the needs of the New Parkland Hospital and its patients. We are also grateful to the leadership team at the Parkland Health and Hospital System for providing strategic guidance regarding how to best maximize the platform’s usefulness to Parkland and to the Parkland IT team for helping with various aspects of the platform build and integration with the EHR. 

 We would also like to express our deep gratitude to Dr. Manjula Julka and Dr. George Oliver, our clinician supervisors, for their patient guidance, enthusiastic encouragement, and useful critiques of this work. We would also like to thank Dr. Shelley Chang, for her providing her clinical expertise, advice, and assistance. Our grateful thanks are also extended to the PCCI leadership team for their support and encouragement throughout this journey. Finally, we would like to thank everyone on the \textit{Isthmus} team for making \textit{Isthmus} a success. 
\section{Disclaimer}
Trademark products mentioned in this paper are properties of respective trademark owners. 

\bibliographystyle{unsrt}  


\end{document}